\documentclass{bmvc2k}
\usepackage{verbatim}
\usepackage{float}

\title{TextCohesion:  Detecting Text for Arbitrary Shapes}

\addauthor{Weijia Wu$^*$}{weijia_wu@yeah.net}{1}
\addauthor{Jici Xing$^*$}{jicixing@gmail.com}{2}
\addauthor{Hong Zhou$\dagger$}{zhouhong_zju@126.com}{1}
\addinstitution{
 Zhejiang University, Key Laboratory for Biomedical Engineering of Ministry, HangZhou, China
}
\addinstitution{
  Zhengzhou University, Industrial Technology Research Institute,\\
  ZhengZhou China
}

\runninghead{Student, Prof, Collaborator}{BMVC Author Guidelines}

\def\eg{\emph{e.g}\bmvaOneDot}

\begin{document}
	
	\maketitle

\begin{abstract}
In this paper, we propose a pixel-wise method named TextCohesion for scene text detection, which splits a text instance into five key components: a Text Skeleton and four Directional Pixel Regions. These components are easier to handle than the entire text instance. A confidence scoring mechanism is designed to filter characters that are similar to text. Our method can integrate text contexts intensively when backgrounds are complex. Experiments on two curved challenging benchmarks demonstrate that TextCohesion outperforms state-of-the-art methods, achieving the F-measure of {\bfseries84.6\%} on Total-Text and {\bfseries86.3\%} on SCUT-CTW1500.

\end{abstract}

\section{Introduction}
\label{sec:intro}
Detecting text in the wild is a fundamental computer vision task which directly determines the subsequent recognition results. Many applications in the real world depend on accurate text detection, such as photo translation \cite{yi2014scene} and autonomous driving \cite{zhu2018cascaded}. Now, horizontal \cite{tian2016detecting,wu2017deep,zhu2017deep}  based methods no longer meet our requirements meanwhile multi-oriented \cite{he2016deep,liu2017deep,liao2017textboxes,zhou2017east,hu2017wordsup,deng2018pixellink} or more flexible pixel-wise detectors \cite{long2018textsnake,zhu2018textmountain,li2018shape} have led to the mainstream. Although these methods are now good enough to be deployed in industrial products, to precisely locate text instances especially captured by mobile sensors is still a big challenge because of arbitrary angles, shapes, and complex backgrounds.

\begin{figure}
\begin{tabular}{ccc}
\bmvaHangBox{\fbox{\includegraphics[width=3.7cm]{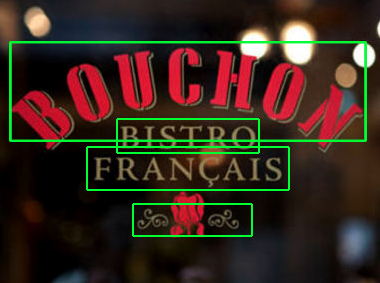}}}&
\bmvaHangBox{\fbox{\includegraphics[width=3.7cm]{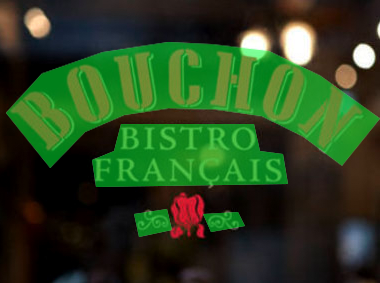}}}&
\bmvaHangBox{\fbox{\includegraphics[width=3.7cm]{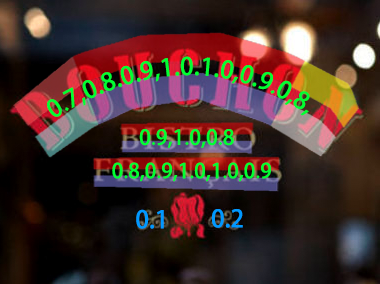}}}\\
(a)&(b)&(c)
\end{tabular}
\caption{(a) Regression-based methods suffer from a fixed shape. (b) Pixel-based methods may not separate from the very adjacent boundaries. (c) TextChohesion.
 Pixel-based and Regression-based methods may face the false positive dilemma. 
}
\label{relatedwork}
\end{figure}

{\bfseries The first challenge is text instances with irregular shapes.} Different from other common objects, the shape of a text instance cannot be accurately described by a horizontal box or an oriented quadrilateral. Some typical methods (\eg{ EAST \cite{zhou2017east}, TextBox++, \cite{liao2018textboxes++}}) perform well in common benchmarks (\eg{ ICDAR 2013 \cite{karatzas2013icdar} ICDAR 2015 \cite{karatzas2015icdar}}) but these regression-based methods may not be gratified in curved text challenges (\eg{Total-Text \cite{ch2017total}, SCUT-CTW1500 \cite{yuliang2017detecting}}) as shown in Figure \ref{relatedwork} (a). 

{\bfseries The second challenge is to cut open text boundaries.} Although pixel-wise methods do not suffer from a fixed shape, they may still fail to separate text areas with very adjacent edges, which shows in Figure \ref{relatedwork} (b).

{\bfseries The third challenge is that text candidates may face the false positives dilemma}\cite{xie2018scene}, because of lacking context information. It can be seen that characters like the bottom of Figure \ref{relatedwork} may confuse detectors. Some symbols or characters that are similar to text may be misclassified.

\begin{figure}[H]
\begin{tabular}{cccc}

\vspace{0cm} 
\setlength{\abovecaptionskip}{5cm} 
\setlength{\belowcaptionskip}{-5cm} 

\bmvaHangBox{\fbox{\includegraphics[width=2.55cm]{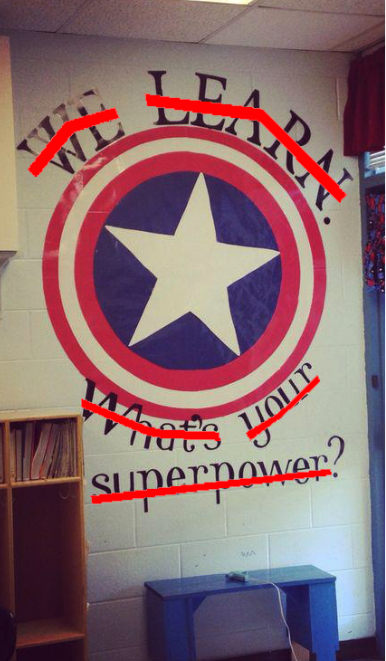}}}
&
\bmvaHangBox{\fbox{\includegraphics[width=2.55cm]{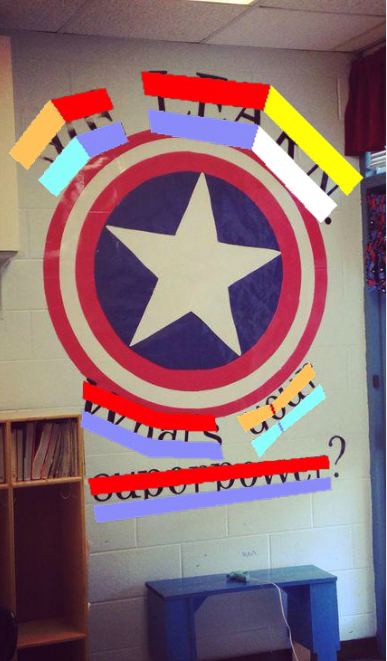}}}&
\bmvaHangBox{\fbox{\includegraphics[width=2.55cm]{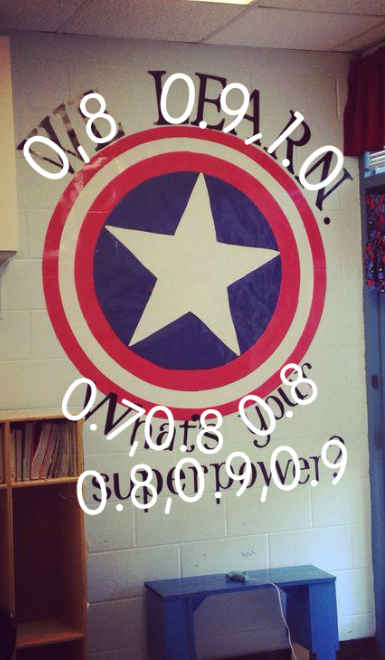}}}&
\bmvaHangBox{\fbox{\includegraphics[width=2.55cm]{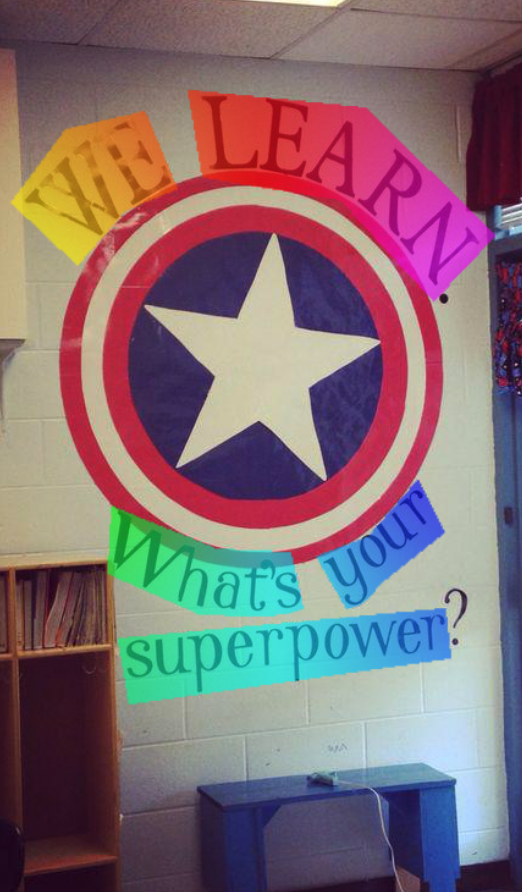}}}\\
(a)&(b)&(c)&(d)
\end{tabular}
\caption{(a) Text Skeleton. (b) Directional Pixel Region. (c) Confidence Scoring Mechanism. The average confidence score of TS is used to filter out false positives. (d) Prediction. Every text instance is constituted of TS and DPR.
}
\label{method}
\end{figure}

To overcome these challenges related above, we propose a novel method, called TextCohesion.
As shown in Figure \ref{method}, our method treats a text instance as a combination of a Text Skeleton and four Directional Pixel Regions where the previous one roughly represents the shape, size while the latter is responsible for refining the text area and edges from four directions. Notably, a pixel belongs to more than one Directional Pixel Regions (\eg{ up, left}), which means the pixel has more chances to be found. Further, the average confidence score of a Text Skeleton that is greater than a defined threshold (\eg{ 0.5}) is considered as a candidate.

The contributions of this paper contain three folds:
\begin{itemize}
    \setlength{\itemsep}{3pt}
    \setlength{\parsep}{3pt}
    \setlength{\parskip}{3pt}
    \item We propose a novel method TextCohesion with three helpers: Text Skeleton, Directional Pixel Region and Confidence Scoring Mechanism to make predictions, which outperforms state-of-the-art methods on curved text benchmarks. 
    \item The proposed method works well for all shapes of text.
    \item The proposed method can further filter out symbols or characters that are similar to text.
\end{itemize}
\section{Related Works}
Detecting text in the wild has been widely studied in the past few years. Before deep learning era, most detectors adopt Connected Components Analysis  \cite{epshtein2010detecting} \cite{huang2013text} \cite{jain1998automatic} \cite{yao2012detecting} \cite{yi2011text} \cite{yin2014robust} or Sliding Window based classification \cite{coates2011text} \cite{lee2011adaboost} \cite{wang2011end} \cite{wang2012end}. 

Now detectors are mainly based on deep neural networks. There are two main trends in the field of text detection: regression-based and pixel-based. Inspired by the promising of object detection architectures such as Faster R-CNN \cite{Ren2015Faster} and SSD \cite{liu2016ssd}, a bunch of regression-based detectors were proposed, which simply regress the coordinates of bounding boxes of candidates as the final prediction. TextBoxes \cite{liao2017textboxes} adopts SSD and adjusts the default box to relatively long shape to match text instances. By modifying Faster R-CNN, Rotation Region Proposal Networks \cite{ma2018arbitrary} inserts the rotation branch to fit the oriented shapes of text in natural images. These methods can achieve satisfying performance on horizontal or multi-oriented text areas. However, they may suffer from the shape of the bounding box even with rotations.
Mainstream pixel-wise methods drew inspirations from the fully convolutional network (FCN) \cite{long2015fully} which removes all fully-connected layers and is widely used to generate semantic segmentation map. Convolution transpose operation then helps the shirked feature restore its original size. TextSnake \cite{long2018textsnake} treats a text instance as a sequence of ordered, overlapping disks centred at symmetric ordered, each of which is associated with potentially variable radius and orientations. It made significant progress on curved text benchmarks. TexeField \cite{xu2019textfield} learns a direction field pointing away from the nearest text boundary to each text point. An image of two-dimensional vectors represents the direction field. SPCNET \cite{xie2018scene}, based on Feature Pyramid Network \cite{Lin2017Feature} and Mask R-CNN \cite{he2017mask}, inserts Text Context Module and Re-Score mechanism to leave the lack of context information clues and inaccurate classification score.

    \section{Methodology}
This section presents details of the TextCohesion. First, we illustrate the feature extractor and how we use of features. Next, we describe the TS, DPR and Confidence Scoring Mechanism of this method. Finally, we manifest how to generate the corresponding label and training objectives.
            \begin{figure*}[!htbp]
            \begin{center}
            \fbox{
            \includegraphics[width=12cm,scale=0.2]{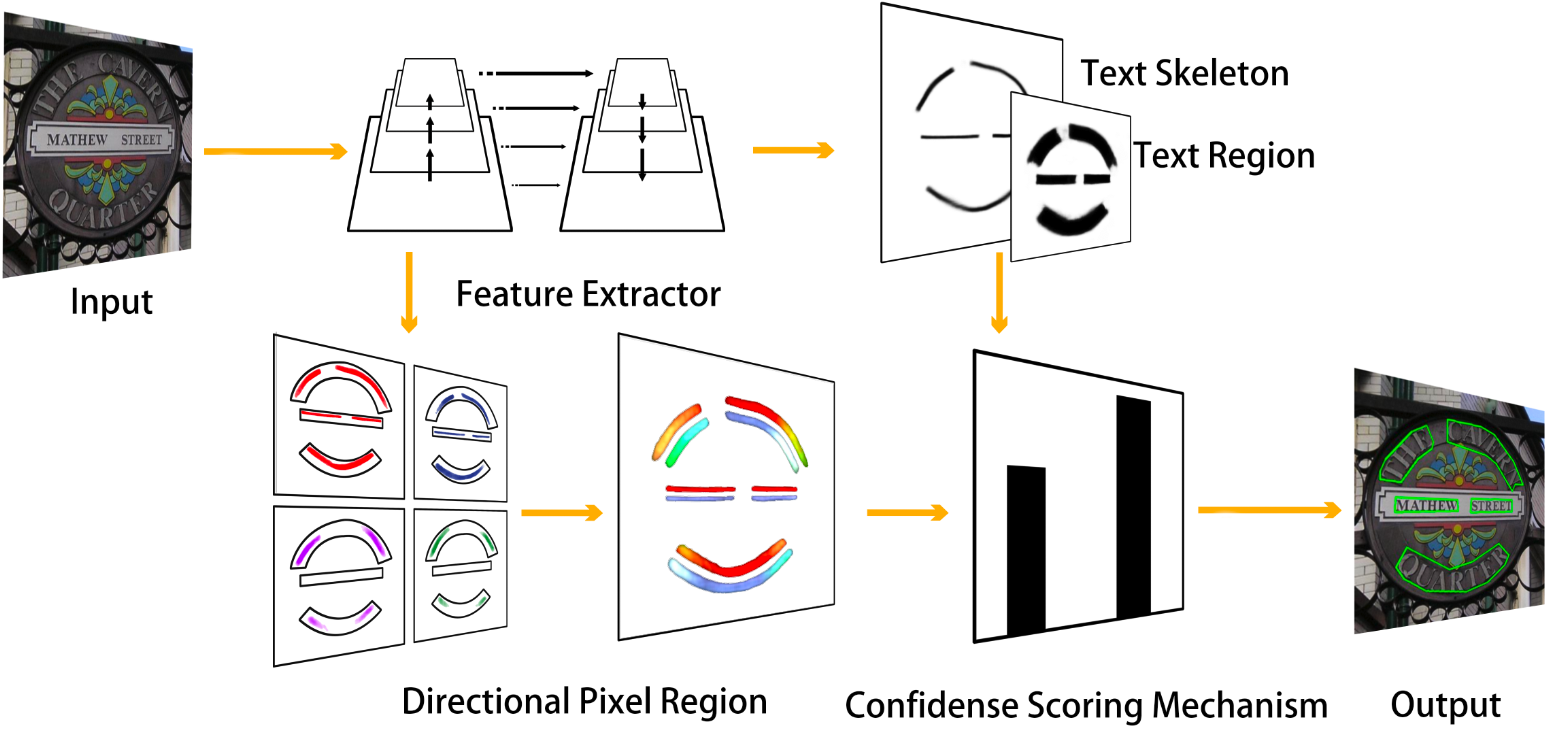}
            }
            \end{center}
               \caption{The pipeline of TextCohesion. To detect irregular text instances, we predict Text Skeleton(TS) and Directional Pixel Regions(DPRs). The post-processing links TS and DPRs to reconstruct the text. All TS are verified by Confidence Scoring Mechanism.}
            \label{ppl}
            \end{figure*}
\subsection{Pipeline}
We introduce Text Skeleton (TS) and Directional Pixel Region (DPR) to precisely capture text instances. For TS, we use a line linked by several dots (\eg{15}) to roughly represent the text instance. Then every DPR is split into several cells by dots of TS. The tangent value between two adjacent dots determines which corresponding cell falls into. Text Region (TR) is a mask that restricts the bounds of TS. Afterwards, the confidence scoring is applied to filter out false positives. Finally, the remaining TS, TR, and DPRs are combined to form the text. The whole process presents in Figure \ref{ppl}.
\subsection{Feature Extractor}
\begin{figure*}[!ht]
            \begin{center}
            \fbox{
            \includegraphics[width=12cm,scale=0.2]{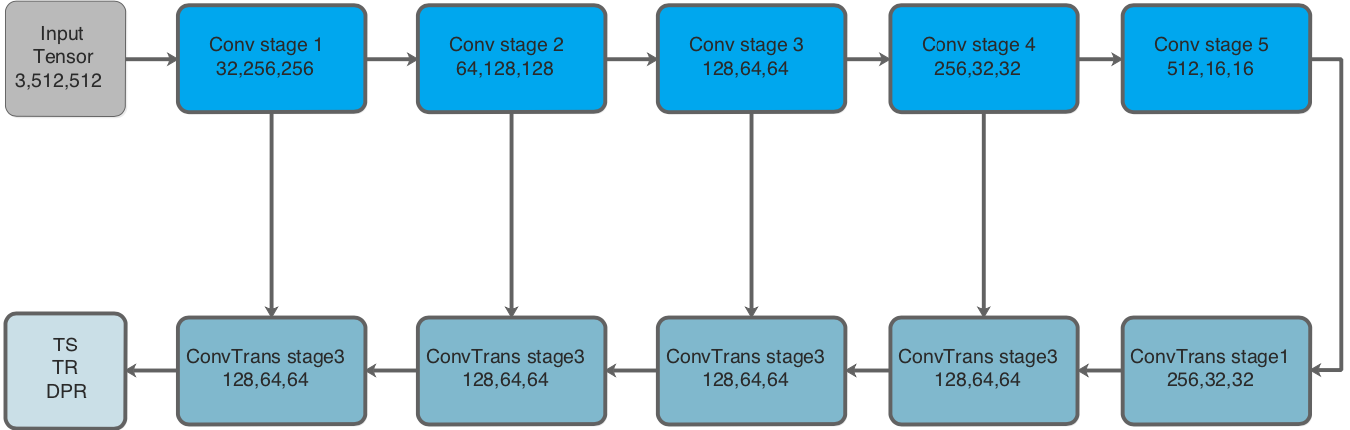}
            }
            \end{center}
               \caption{Feature Extractor}
            \label{fe}
            \end{figure*}

 We chose VGG16 \cite{simonyan2014very} as our feature extractor for a fair comparison. Inspired by FPN \cite{Lin2017Feature}, lateral connections are also inserted to enrich features, and the feature extractor shows in Figure \ref{fe}. At the first stage, images are downsampled to the multilevel features. Secondly, features are gradually upsampled to the original size and mixed with the corresponding output of the previous stage. Then several maps are generated to represent TS, DPR, and TR.
    \subsection{Text Skeleton}
As shown in Figure \ref{lb1} (a), we use TS to roughly represent text candidates. Specifically, the dots in TS are treated as a series of starting points to future searching the corresponding regions of interest. Moreover, TS is also used to filter out false positives. Compared with the entire text instance, TS is less confused by adjacent boundaries, easier to locate, and can represent the shape of the original text approximately. Therefore, we consider every TS as a candidate.
    \subsection{Directional Pixel Region}
\begin{figure}
\begin{tabular}{cccc}
\bmvaHangBox{\fbox{\includegraphics[width=2.65cm]{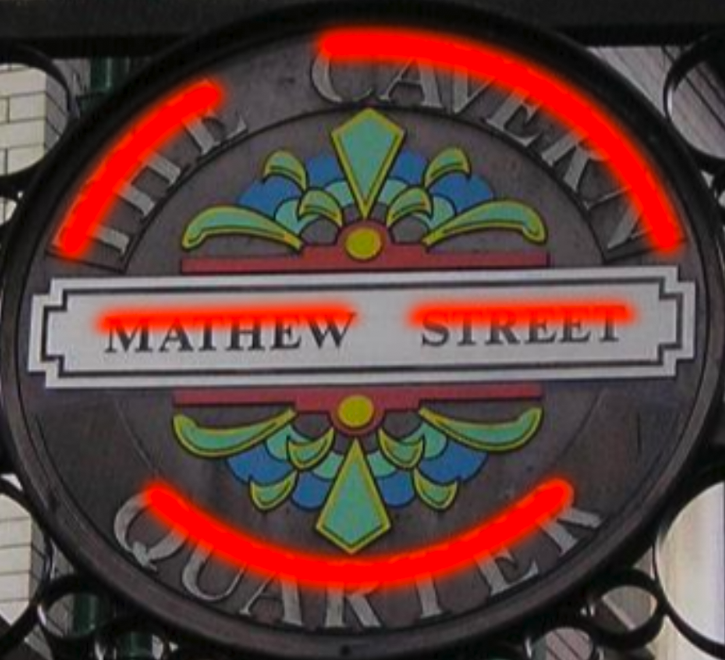}}}
&
\bmvaHangBox{\fbox{\includegraphics[width=2.65cm]{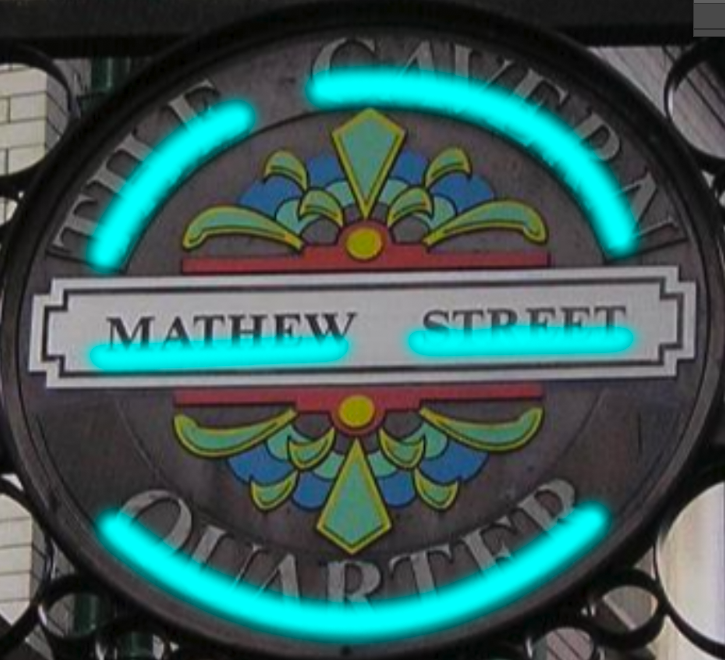}}}&
\bmvaHangBox{\fbox{\includegraphics[width=2.65cm]{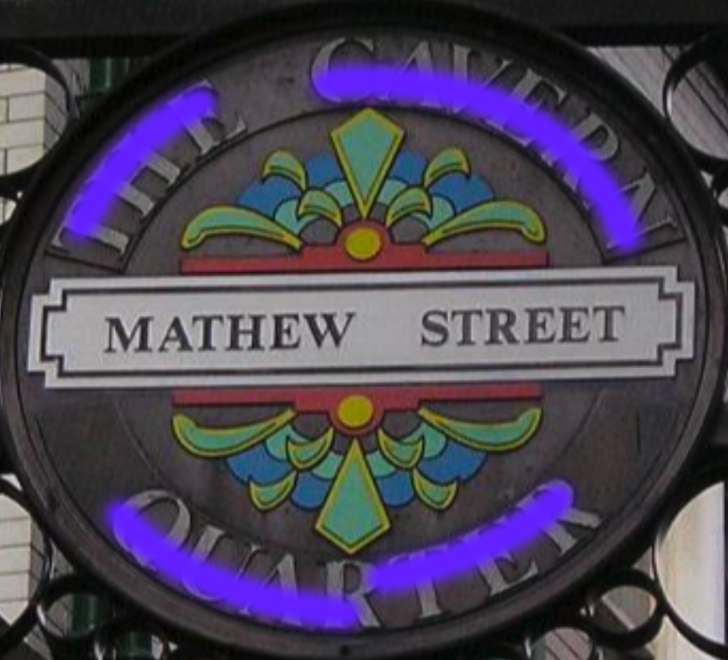}}}&
\bmvaHangBox{\fbox{\includegraphics[width=2.65cm]{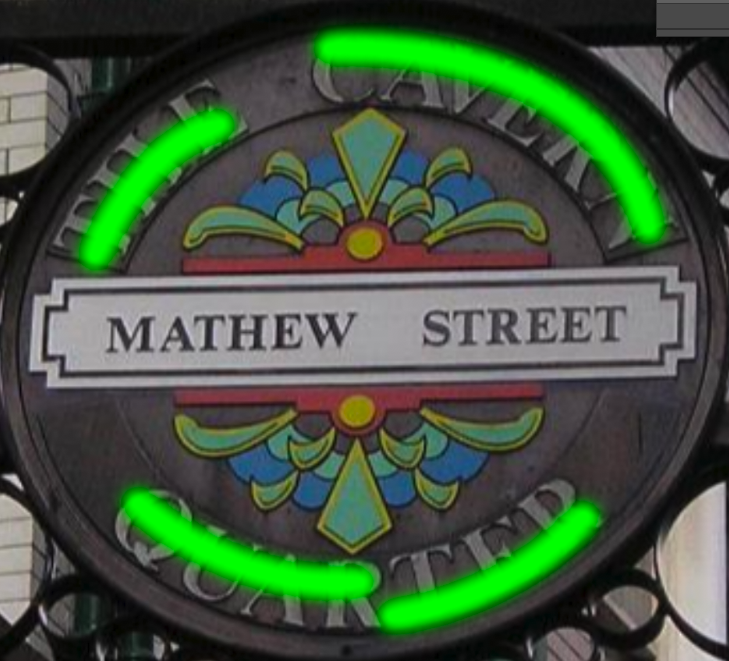}}}\\
(a)&(b)&(c)&(d)
\end{tabular}
\caption{(a) up region (b) down region (c) left region (d) right region.}
\label{four}
\end{figure}

Pixels that in text instance but not in TS are categorized into four regions as shown in Figure \ref{four}. 
DPRs are used to segment the edges elaborately. Notably, overlapping happens in more than one region (\eg{ up, right}). All information caught by four directions, in general, would be twice the original image approximately. A pixel would have a higher confidence score when it is ensured by more directions. It makes our method more robust, especially in complex background.
     \subsection{Confidence Scoring Mechanism}
To filter out false positives, we consider values in TS as the probability of the actual text, instead of simple flags. They are calculated by the following equation:
\begin{align}
    \begin{cases} TP,&\frac{\sum _{i=1}^{n}Ts\left( i\right) }{n}>\gamma \\FP,&Other\end{cases} 
\end{align}
where $n$ is the total number of pixels in TS and $\gamma$ is the threshold value to filter out those TS with low confidence score. Note that the Confidence Scoring Masochism is easy to transfer on other detectors.
        \subsection{Label Generation}
        \subsubsection{Label of Text Skeleton}

\begin{figure}[!htbp]
\begin{tabular}{cc}
\centering
\bmvaHangBox{\fbox{\includegraphics[width=5.8cm]{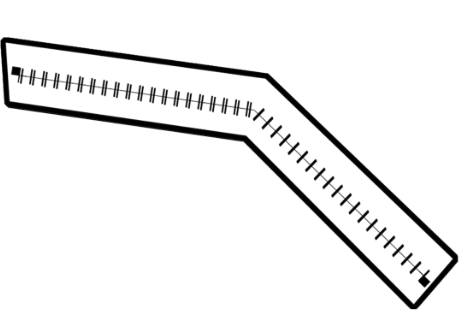}}}&
\bmvaHangBox{\fbox{\includegraphics[width=5.8cm]{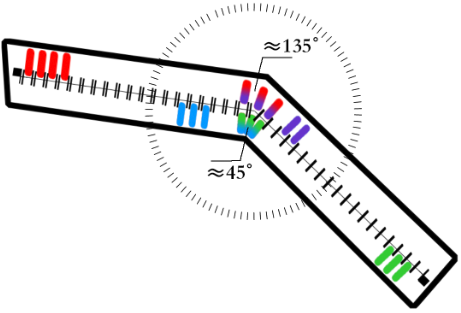}}}\\
(a)&(b)
\end{tabular}
\caption{(a) Text Skeleton is divided by some dots. (b) Directional Pixel Regions are filled by the tangent angle of dots.}
\label{lb1}
\end{figure}

TS is a line expanded by the average values of the sampled points of two longest edges in a text area, which represents condensed information of a text instance. The appearance of TS  presents in Figure \ref{lb1} (a). For a text instance $t$ represented by a group of vertexes $\left\{ v1,v2,v3, \dots, vn \right\}$ in a certain order, we firstly pick some vertexes that would change the area considerably $\left\{ v1,v2,v3, \dots, v6 \right\}$. Then linking the vertexes to get a connected graph and compute the lengths of all edges. The two shortest edges are discarded, and the remaining two boundaries are what we need. Finally, the line between the two boundaries is TS.

    \subsubsection{Label of Directional Pixel Region}
As shown in Figure \ref{lb1} (b), pixels that in text instance but not in TS  are categorized into 4 regions  which are defined by the following equation:
\begin{align}{\begin{cases}up, &\left[ {\sum _{i=1}^{n}}  {\tan(-30) < \tan \left( s_{i}-s_{i-1}\right)   <\tan(30)}  \right]\cap \left( y_{i},y_{i-1}\right) <TS  \\
down,&\left[ {\sum _{i=1}^{n}}  {\tan(-30) < \tan \left( s_{i}-s_{i-1}\right)   < \tan(30)}  \right]\cap \left( y_{i},y_{i-1}\right) >TS
\\ left,&\left[ {\sum _{i=1}^{n}}  {\tan(-60) < \tan \left( s_{i}-s_{i-1}\right)   < \tan(60)}  \right]\cap \left( x_{i},x_{i-1}\right)<TS\\ right,&\left[ {\sum _{i=1}^{n}}  {\tan(-60) < \tan \left( s_{i}-s_{i-1}\right)   < \tan(60)}  \right]\cap \left( x_{i},x_{i-1}\right)
>TS\\all, &other \end{cases} }
\label{eq4d}
\end{align}

where $\left\{ s_{1},s_{2},\dots, s_{i-1} \right\}$ are dots of $TS$ , $(x,y)$ is a coordinate of pixel and $n$ is the sum of pixels. In particular, a portion of dots with a tangent value near to 45 or 135 degrees is difficult to define its regions, so they are occupied by more than one regions (\eg{ top, left}). Specifically, these dots with the tangent value among 30 to 60 and -30 to -60 are hard to define, so all of them are considered to belong to two adjacent regions.

\subsection{Training Objectives}
The proposed model is trained end-to-end, with the following loss functions as the objectives:
\begin{align}
  L =\lambda L _{TS}+L _{DPR}+L _{TF}+L_{TR}
  \label{loss}
\end{align}
Where $\lambda$ is set to 3.0 experimentally, and $L _{TS}$ is a self-adjust cross entropy loss \cite{deng2018pixellink}, because putting the same weight on all positive pixel is unfair, in which case the large instance contributes greater loss while the smaller one little. The total loss should treat all samples equally regardless of their size.
\begin{align}
  L _{TS} = \frac{B}{S_{i}}\sum_{n=1}^NCrossEntropy(TS_{i},\widehat{TS_{i}})
  \label{ts}
\end{align}
where $B$ is the sum of text instances in an image, $S _{i}$ represent the size of $ith$  instance. $TS_{i},\widehat{TS_{i}}$ are the predicted $ith$ pixel belonging to TS and the corresponding ground truth respectively. 
\begin{align}
    L_{DPR} = \sum_{n=1}^4\sum_{i\in DPR_{n}}SmothL1(DPR_{i},\widehat{DPR_{i}})
\end{align}
where $n$ represents the $ith$ $DPR$. $DPR_{i},\widehat{DPR_{i}}$ represent the $ith$ pixel falling into and its ground truth. We use Smooth L1 loss \cite{Ren2015Faster} in case of outlier effects.

For $L_{TF},L_{TR}$, we also choose cross entropy loss. $L_{TF}$ is responsible for penalizing false positives and $L_{TR}$ is used to prevent points out of boundaries.

\section{Experiments}
In this section, we evaluate the proposed method on curved challenging benchmarks for scene text detection and compare it with other algorithms.

SynthText \cite{gupta2016synthetic} is a large scale dataset that contains about 800K synthetic images. These images are created by blending natural images with text rendered with random fonts, sizes, colours, and orientations. Thus these images are quite realistic. We use this dataset to pre-train our model.

Total-Text \cite{ch2017total} is a word-level based English curve text dataset which is split into training and testing sets with 1255 and 300 images respectively.

SCUT-CTW1500 \cite{yuliang2017detecting} is another dataset mainly consisting of curved text. It consists of 1000 training images and 500 test images.
\subsection{Data Augmentation}
Images are randomly rotated, cropped and mirrored. After that, colour, and lightness are randomly adjusted. We ensure that the text on the augmented images are still legible if they are legible before augmentation.
\subsection{Implementation Details}
Our method is implemented in Pytorch \cite{paszke2017automatic}. The network is pre-trained on SynthText for two epoch and fine-tuned on other datasets. We adopt Adam optimizer as our learning rate scheme. During the pre-training phase, the learning rate is fixed to 0.001. During the fine-tuning stage, the learning rate is initially set to 0.0001 and decays with a rate of 0.94 every 10000 insertions. All the experiments are conducted on a regular workstation(CPU: Intel(R) Core(TM) i7-7800X CPU @ 3.50GHz; GPU: GTX 1080). We train our model with a batch of 4 on one GPU.
\subsection{Experiment Results}
Experiment on Total-Text Fine-tuning stops at 35 epochs. Thresholds for $T _{TS}, T _{TR}, T_{DPR}$ are set to 0.54, 0.2, 0.1 respectively.
In training and testing, all images are resized to 512 * 512. We choose the following models for comparison which can be found in Table \ref{total}.
\begin{table}[H]
\begin{center}
\begin{tabular}{|l|c|c|c|}
\hline
Method & Precision & Recall & F-measure \\
\hline\hline
DeconvNet \cite{neumann2010method}& 33.0 &40.0 & 36.0 \\
EAST \cite{zhou2017east}& 50.0&36.2&42.0 \\
TextSnake  \cite{long2018textsnake}&82.7&74.5&78.4\\
TextField \cite{xu2019textfield}&79.9&81.2&80.6\\
CSE \cite{Liu2019Towards} &81.4&79.7&80.2\\
PSENet-1s \cite{li2018shape} &84.02&77.96&80.87\\
SPCNET\cite{xie2018scene} &82.8&\bfseries83.0&82.9\\
CRAFT \cite{baek2019character} &87.6&79.9&83.6\\
Ours & \bfseries88.1 &81.4&\bfseries84.6\\
\hline
\end{tabular}
\end{center}
\caption{Experimental results on Total-Text.}
\label{total}
\end{table}
On SCUT-CTW1500, we use the same settings as Total-Text except for the values of $T _{TS}$ is set to 0.29. The comparison can be found in Table \ref{ctw1500}.
\begin{table}[H]
\begin{center}
\begin{tabular}{|l|c|c|c|}
\hline
Method & Precision & Recall & F-measure \\
\hline\hline
SegLInk \cite{Shi2017Detecting} & 42.3 &40.0 & 40.8 \\
EAST \cite{zhou2017east} & 78.7&49.1&60.4 \\
DPMNet \cite{liu2017deep} &69.9&56.0&62.2\\
CTD \cite{Noh2015Learning}&74.3&65.2&69.5\\
CTD+TLOC \cite{Noh2015Learning}&77.4&69.8&73.4\\
TextSnake \cite{long2018textsnake}&67.9&\bfseries85.3&75.6\\
PSENet-1s \cite{li2018shape} &84.8&79.7&82.2\\
TextMountain \cite{zhu2018textmountain} & 82.9&83.4&83.2\\
PAN MASK R-CNN \cite{huang2019mask} &86.8&83.2&85.0\\
Ours & \bfseries 88.0 &84.7&\bfseries86.3\\
\hline
\end{tabular}
\end{center}
\caption{Experimental results on SCUT-CTW1500.}
\label{ctw1500}
\end{table}

Our method achieves the highest F-measure in these benchmarks. It turns out that TS may be an appropriate representation rather than dealing directly with the entire text instance. After candidates selection, a text instance is initialized as its corresponding TS, and then gradually diffuses outward along the DPRs belonging to this TS. In this process, pixels belonging to a specific DPR will be firstly searched in that direction (\eg{The up region will be first searched up along TS}), and then there will be other chances to be supplement from different searching paths (\eg{The up region will be supplemented by searching the left and right regions}). In other words, the direction of a pixel is not unique, and a text instance will have multiple opportunities to recover completely.

\subsection{Transferable Confidence Scoring Mechanism}
\label{sec:exptcl}
To test the effectiveness of Confidence Scoring Mechanism, we implement it on SCUT-CTW1500. 
\begin{table}[H]
\begin{center}
\begin{tabular}{|l|c|c|c|}
\hline
Method & Precision & Recall & F-measure \\
\hline\hline
TextCohesion with no scoring mechanism & 82.3 &\bfseries84.9 &83.5 \\
TextCohesion with  scoring mechanism & \bfseries88.0&84.7&\bfseries86.3 \\
\hline
\end{tabular}
\end{center}
\caption{Effectiveness of Confidence Scoring Mechanism on SCUT-CTW1500.}
\label{comp}
\end{table}
In Table \ref{comp}, the precision is improved significantly when Confidence Scoring Mechanism is inserted. We argue that this might help other detectors to resolve the precision related problems.
\subsection{Conclusion and Future Work}
\begin{figure}[!htbp]
\begin{tabular}{ccc}
\bmvaHangBox{\fbox{\includegraphics[width=12cm]{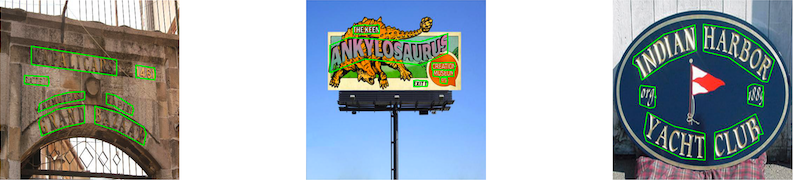}}}\\

(a)\\

\bmvaHangBox{\fbox{\includegraphics[width=12cm]{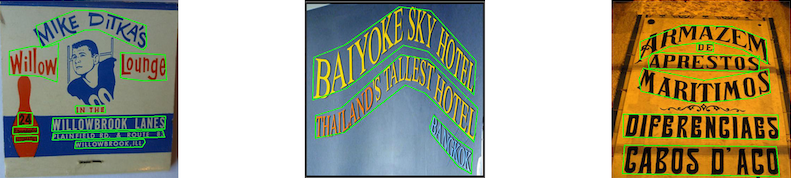}}}\\

(b)

\end{tabular}
\caption{(a) Total-Text result (b)SCUT-CTW1500 result 
}
\label{resultfig}
\end{figure}

In this paper, we introduce a novel text detector that outperforms state-of-the-art methods on curved text benchmarks (Figure \ref{resultfig}). The novel proposed Directional Pixel Region and Text Skeleton with a scoring mechanism may be the key factors to make it sensible on text with arbitrary shapes, dedicated boundaries, and false positive dilemma. In future research, we would explore an end-to-end detection with the recognition system for text with arbitrary shapes.

\bibliography{egbib}
\end{document}